%% file: main.tex
\documentclass[sigconf,10pt]{acmart}
\pdfoutput=1 

\usepackage{svg}
\usepackage[T1]{fontenc}
\usepackage{algorithm2e}
	\RestyleAlgo{algoruled}
\usepackage{amsmath}
\usepackage{amsfonts}
\usepackage{colortbl}
\usepackage{fp}
\usepackage{graphicx}
\usepackage{hyperref}
\usepackage{multirow}
\usepackage{soul}
\usepackage{subcaption}
\usepackage{textcomp}
\usepackage{url}
\usepackage{xcolor}
\usepackage{xspace}

\makeatother

\DeclareMathOperator*{\argmax}{argmax}

\definecolor{Gray}{gray}{0.9}
\newcolumntype{g}{>{\columncolor{Gray}}c}


\begin{document}
\title{Transfer-Learning-Based Autotuning Using Gaussian Copula}

\copyrightyear{2023}
\acmYear{2023}
\setcopyright{licensedusgovmixed}\acmConference[ICS '23]{2023 International Conference on Supercomputing}{June 21--23, 2023}{Orlando, FL, USA}
\acmBooktitle{2023 International Conference on Supercomputing (ICS '23), June 21--23, 2023, Orlando, FL, USA}
\acmPrice{15.00}
\acmDOI{10.1145/3577193.3593712}
\acmISBN{979-8-4007-0056-9/23/06}

\author{Thomas Randall} \affiliation{  \institution{Clemson University} \country{USA}} \email{tlranda@clemson.edu} \authornote{Authors contributed equally to this work.}

\author{Jaehoon Koo} \authornotemark[1] \affiliation{\institution{Hanyang University} \country{Republic of Korea}} \email{jaehoonkoo@hanyang.ac.kr}

\author{Brice Videau} \affiliation{\institution{Argonne National Laboratory} \country{USA}} \email{bvideau@anl.gov}

\author{Michael Kruse} \affiliation{\institution{Argonne National Laboratory} \country{USA}} \email{michael.kruse@anl.gov}

\author{Xingfu Wu} \affiliation{\institution{Argonne National Laboratory} \country{USA}} \email{xingfu.wu@anl.gov}

\author{Paul Hovland} \affiliation{\institution{Argonne National Laboratory} \country{USA}} \email{hovland@anl.gov}

\author{Mary Hall} \affiliation{\institution{University of Utah} \country{USA}} \email{mhall@cs.utah.edu}

\author{Rong Ge} \affiliation{\institution{Clemson University} \country{USA}} \email{rge@clemson.edu}

\author{Prasanna Balaprakash} \affiliation{\institution{Oak Ridge National Laboratory} \country{USA}} \email{pbalapra@ornl.gov}

\renewcommand{\shortauthors}{Randall et al.}

\authornotemark[1]

\newcommand{\pasteKeywords}{
Transfer Learning, Autotuning, Few-Shot Learning, Gaussian Copula
}

\input{sections/abstract.tex}
\maketitle

\section{Introduction} \label{sec:intro} \input{sections/intro.tex}
\vspace{-1em}
\section{Background} \label{sec:background} \input{sections/background.tex}
\section{Proposed Framework} \label{sec:methodology} \input{sections/methodology.tex}
\section{Experiment Design} \label{sec:design} \input{sections/design.tex}
\vspace{-1em}
\section{Results} \label{sec:results} \input{sections/experiments.tex}
\vspace{-1em}
\section{Related Work} \label{sec:relwork} \input{sections/relwork.tex}
\section{Conclusions} \label{sec:conclusions} \input{sections/conclusions.tex}

\section*{Acknowledgment}
This research was partially supported by the Exascale Computing Project (17-SC-20-SC), a collaborative effort of the U.S. Department of Energy Office of Science and the National Nuclear Security Administration, and by U.S. National Science Foundation 
 under Grants CCF-1942182. This material is based upon work supported by the U.S. Department of Energy, Office of Science, under contract number DE-AC02-06CH11357. 

\balance
\bibliographystyle{ACM-Reference-Format}
\bibliography{main}

\end{document}

%% file: sections/abstract.tex
\begin{abstract}
As diverse high-performance computing (HPC) systems are built, many opportunities arise for applications to solve larger problems than ever before. 
Given the significantly increased complexity of these HPC systems and application tuning, empirical performance tuning, such as autotuning, has emerged as a promising approach in recent years. Despite its effectiveness, autotuning is often a computationally expensive approach.
Transfer learning (TL)-based autotuning seeks to address this issue by leveraging the data from prior tuning.
Current TL methods for autotuning spend significant time modeling the relationship between parameter configurations and performance, which is ineffective for few-shot (that is, few empirical evaluations) tuning on new tasks.
We introduce the first generative TL-based autotuning approach based on the Gaussian copula (GC) to model the high-performing regions of the search space from prior data and then generate high-performing configurations for new tasks.
This allows a sampling-based approach that maximizes few-shot performance and provides the first probabilistic estimation of the few-shot budget for effective TL-based autotuning.
We compare our generative TL approach with state-of-the-art autotuning techniques on several benchmarks.
We find that the GC is capable of achieving 64.37\% of peak few-shot performance in its first evaluation.
Furthermore, the GC model can determine a few-shot transfer budget that yields up to 33.39$\times$ speedup, a dramatic improvement over the 20.58$\times$ speedup using prior techniques.
\end{abstract}

%% file: sections/intro.tex
The arrival of diverse architectures in high-performance computing (HPC) systems has unlocked many new opportunities and also permits existing applications to push beyond their former limitations.
In order to maximize performance, new and old applications alike will need to be tuned.
Since these applications frequently allow many potential optimizations, searching for the best configurations by hand or with exhaustive enumeration is typically too expensive.
Empirical performance tuning, widely known as autotuning, is a promising approach that evaluates a small subset of parameter configurations of a given kernel or application from a large user-defined search space by running them on the target platform to identify the best-performing configurations.
A sophisticated search algorithm is often employed to intelligently navigate the large search space.
Such autotuning approaches have achieved success in several prior works~\cite{WP23,WK21,ytopt,Ansel2012,GPTune,YouOnlyRunOnce,Roy2021}.

Despite prior successes, however, autotuning has faced adoption challenges for real applications because it is still resource expensive.
Each empirical evaluation involves generating the executable with the parameter configuration and actual execution. 
Even simple kernels may require several hours to tune, while more advanced and complex applications with larger search spaces may require days.
To reduce the computational expense of autotuning, researchers have developed transfer learning (TL) methods to leverage data from related autotuning tasks (e.g., similar input sizes or kernels).
Although the optimum for a kernel changes with input size, high-performing regions in the search space are related across input sizes.
This allows TL in autotuning to tune new input sizes of that kernel efficiently.

Existing TL autotuning methods are ineffective for few-shot, \textit{i.e., a minimal number of empirical evaluations}, as they require many samples for new tasks to model the transfer relationship.
To overcome this issue, we develop a new generative autotuning approach that uses Gaussian copula (GC), a data-efficient statistical model, to enable rapid TL autotuning.
We use GCs to model each configuration parameter's distribution and codependencies.
GCs permit generative tuning via conditional sampling, which restricts sample generation to configurations to satisfy constraints such as high performance for the input size or kernel of interest.
Conditional sampling enhances the explainability of generated configurations and improves the likelihood of success on transferred problems.
We enhance the GC's ability to model the marginal and joint distributions of parameters while mitigating its limitations for autotuning settings.

Our main contributions are as follows:
\vspace{-1em}
\begin{itemize}
	\item A new generative modeling approach based on a data-efficient GC model, which enables few-shot TL based autotuning with a small number of empirical evaluations for new tasks; a generative modeling approach has never been developed or applied for TL autotuning before.
    \item Estimation of success probability for generative modeling to determine the necessary budget to expect quality autotuning results; this is the first work that provides probability estimation for TL autotuning.
    \item We demonstrate new performance insights for Polybench and Exascale Computing Project mini-applications by utilizing few-shot autotuning.
\end{itemize}

Our code is open source and available at \url{https://github.com/tlranda/GC_TLA}.

%% file: sections/background.tex
\subsection{Autotuning}

Autotuning~\cite{balaprakash2018autotuning,WK21,WP23} is a process that efficiently evaluates a number of parameter configurations from a user-defined parameterized kernel or application to optimize a given objective such as performance (e.g., runtime, FLOPS).
Here we provide a walkthrough with the Polybench kernel~\cite{polybench} ``3mm'' as a concrete example of basic autotuning concepts.
The kernel performs dense matrix multiplication with four matrices $A,B,C,D$ such that the output is $(A \times B) \times (C \times D)$.

Autotuning utilizes a finite budget (typically time or number of evaluations) to optimize a relationship $f(c;t) \in \mathbb{R}^d$ between a given parameter configuration $c$, out of all possible configurations $C$, a tuning task $t$, and $d$ objective outputs such that $\argmax_c f(c; t) \; \forall c \in \mathcal{C}$.
Each task $t$ is a specific instance from a set of related tasks $\mathcal{T}$, which may have different configurations for optimum performance.
Each objective $d$ is a real-valued metric that functionally depends on both the task and parameters according to $f(c;t)$.
The exact closed form of $f(c;t)$ is unknown but is assumed to be a complex, nonlinear relationship.

An example task of tuning the 3mm kernel's runtime performance involves $n=10$ parameters in the form of source code annotations that affect loop tile sizes (i.e., 4, 8, 32), loop interchanges (the order loop iterators appear in nested loops), and memory management (the packing used for tile memory structures).
Each evaluation of the objective requires annotating the source with parameter values, then compiling and executing it on the benchmark system to collect timing data, which incurs considerable cost even for small input matrices.
There are 376,320 unique combinations of the ten parameters that define our tuning space for 3mm, which is prohibitively costly to brute-force with empirical searches.
Autotuning uses more intelligent approaches to identify the configurations that achieve optimal performance.

Autotuning must differentiate input scales as different tasks because changing the input scale frequently induces drastic changes in the optimum configuration.
As shown in Table~\ref{tbl:3mm_Sizes}, small sizes require the packed-array technique for matrices $A$ and $E$, but medium-sized inputs do not.
The degree of improvement can also vary between input scales, where small 3mm inputs can gain 1.13$\times$ speedup from autotuning.
However, medium-sized 3mm inputs gain 14.94$\times$ speedup over the respective baselines.

\begin{table}[t]
	\begin{center}
		\caption{Matrix input scales affect speedup and the best configurations for the 3mm kernel.}
        {\scriptsize
		\begin{tabular}{|l|l|l|l|}
		\hline \multirow{2}{*}{} &
			\multicolumn{3}{|c|}{\textbf{Input Scale}}
			\\ \cline{2-4}
		 & 
			Small &
			Medium &
			Large 
			\\ \hline
		\multicolumn{4}{|c|}{\textbf{Input Scale Characteristics}}
			\\ \hline
		Array Dimensions &
			$\le80$ &
			$\le220$ &
			$\le1200$ 
			\\ \hline
		Naive Tera-Ops &
			     0.037 & 
			     4.75 & 
			  2924.24 
			\\ \hline
		Worst Runtime (s) &
			  0.00017 & 
			  0.1096 & 
			  9.8631 
			\\ \hline
		\multicolumn{4}{|c|}{\textbf{Best Configuration Values}} \\ \hline
		Packed Arrays &
			A,E,F &
			F &
			A,B,E 
			\\ \hline
		Loop Interchanges &
			N/A &
			N/A &
			Outer Exchange 
			\\ \hline
		Tile Sizes &
			16, 2048, 4 &
			96, 16, 4 &
			4, 2048, 4 
			\\ \hline
		Speedup Over Default &
			  1.13$\times$ & 
			 14.94$\times$ & 
			 50.50$\times$ 
			\\ \hline
		\end{tabular}
        }
	\label{tbl:3mm_Sizes}
	\end{center}
\end{table}

\vspace{-1em}
\subsection{Transfer Learning in Autotuning} \label{section:background+transfer}

Several search methods have been developed to reduce the number of evaluations required to find the best configuration for autotuning tasks.
They can be classified into model-based and model-free methods.
The former methods learn the relationship between the parameter configurations and the objective function through an incrementally updated surrogate model and leverage it to cheaply evaluate multiple points and minimize the number of actual evaluations.
Examples include Bayesian optimization that employs Gaussian process regression and random forest and their variants.
The latter methods optimize the objective function without such models.
Examples include random search, grid search, genetic algorithms, and Nelder-–Mead.
The key advantage of the model-based methods is that they require significantly fewer evaluations than the model-free methods, especially for large search spaces~\cite{GPTune,ytopt,WP23,WK21}.

TL in autotuning is an emerging approach that leverages data from one autotuning task in related autotuning tasks to improve sample efficiency significantly.
Related autotuning tasks are common in HPC applications, which include tuning different input sizes of the same kernel or application, tuning the same kernel across architectures, and tuning related kernels with the same computational signature.
While the best configurations are often different for different autotuning tasks, TL is particularly effective when the related tasks share similar high-performing characteristics in the search space.
Model-based search methods are promising for TL because the model can be pretrained or bootstrapped with the existing data from related tasks. 

\vspace{-1em}
\subsection{Gaussian Copula}

The generative modeling-based TL approach that we propose is based on the GC, a well-known multivariate probability distribution in statistics literature.
Let us consider a simple autotuning example with three variables: input scale, one tunable parameter, and the performance metric.
After tuning several input scales, we can model the distribution of the values of the three variables independently.
These are referred to as marginal distributions.
The three variables are correlated, however, so we also model their interactions with one another using a joint probability distribution.

Copulas are a class of statistical modeling techniques that decompose a multivariate probability distribution into its marginal distributions and use a separate function to couple those distributions.
This approach allows us to specify the correlation separately via a correlation matrix.
GCs adopt probability integral transform, a technique that can transform any probability distribution into a uniform distribution and vice versa.
GCs use the uniform and normal distribution as the intermediate distribution to model complex joint probability distribution.
This is achieved as follows.
Given the values of the three variables, a covariance matrix that models the correlation among the variables is computed.
A multivariate normal distribution is then defined using the computed covariance matrix with a zero mean vector.
The probability integral transform is applied to convert the marginals of the Gaussian distribution to uniform distributions.
The uniform marginal distributions are then converted into the original distribution using the probability integral transform.
We refer the reader to the work of Masarotto and Varin~\cite{GCMR} for a more detailed mathematical exposition of the statistical model and mechanics.

%% file: sections/methodology.tex
The key idea of our TL approach is to leverage the GC to predict high-performing configurations on related tasks in few-shot autotuning.

\begin{figure}[!h]
	\centering
	\def\svgwidth{\columnwidth}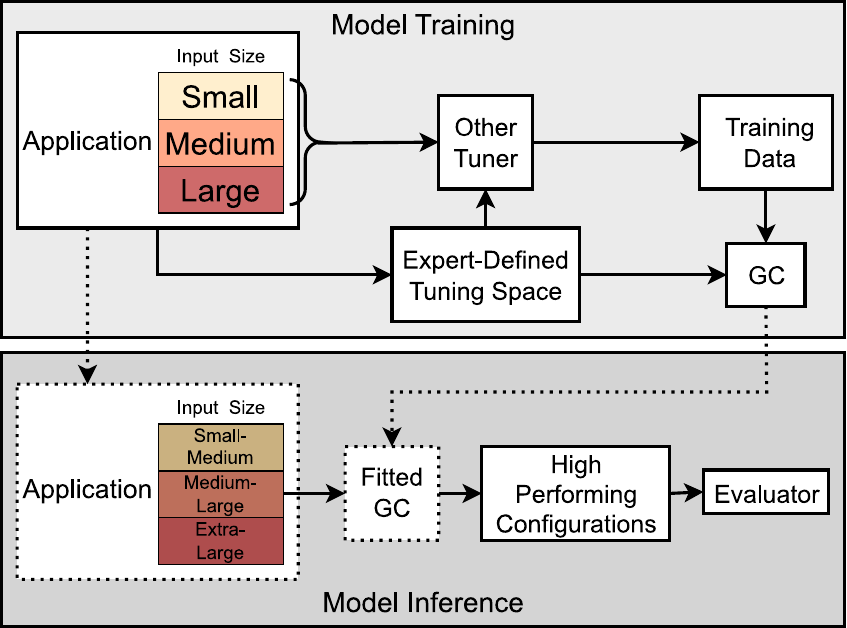
	\caption{\textbf{TL-based Autotuning Framework Using GC.} TOP: Model Training, which uses GC to train fitted models with data collected from source tasks (multiple input sizes of an application) in a human-designed tuning space. \textbf{BOTTOM: Model Inference,} which uses the fitted GC models to propose high-performing configurations for new tasks and evaluates them.}
	\label{fig:overall_approach}
\end{figure}

Our proposed method consists of two phases: model training and model inference, as shown in Figure~\ref{fig:overall_approach}.
Model training uses GC to fit data collected from source tasks in an expert-defined tuning space.
In our work, the source tasks correspond to different input sizes for the same application, and the tuning space is specified via application source code annotation and predefined parameter values.
The source tasks, tuning space, and source input sizes are presented to an existing autotuner~\cite{ytopt,Roy2021,Ananta2011,Chen2015,ParEGO,Tuneful,EfficientCompilerAutotuningViaBO} with a fixed evaluation budget to collect a small, quality training dataset of empirical performance.
Model inference uses the fitted GC model to propose high-performing configurations for new tasks, which are then empirically evaluated.
We discuss the modules in greater detail in the remainder of this section.

\vspace{-1em}
\subsection{Model Training}

Autotuning problems require experts to delineate key high-level features.
The GC also requires several interventions to permit its general usage in autotuning and to improve its utility as a few-shot TL autotuner.

\subsubsection{GC for Autotuning}

We make several adaptations for GC to generalize it for the autotuning problem.

\paragraph{Variable Preprocessing}
Standard GCs model real variables but do not model mixed-integer (discrete, integer, and categorical) variables.
To address this issue, we adopt a new GC approach proposed for synthetic data generation~\cite{Sdv}.
In this GC approach, numeric variables (real or integer) are modeled by truncated Gaussian distributions, and categories are reordered by their frequency in the fitting data.
The GC also reduces the bias from distribution shape by converting all variable distributions to standard normal distribution before computing covariance.

\paragraph{GC as an Autotuner}

GC can be used as an autotuner given a dataset of observed configurations in a defined tuning space.
Each tunable parameter in an autotuning space can be represented by a marginal variable in the GC model; the combination of parameter interactions can be described through the joint model of the GC, permitting representation of distributions throughout a tuning space.
The resulting fitted model identifies appropriate marginal and joint distributions.
The fitted model can generate new configurations through the GC's probability integral transform, which statistically resembles the training data's observed marginal and joint behaviors.
Any configurations a GCs generates can be empirically evaluated to determine its fitness without iterative feedback.

\subsubsection{GC Model Fitting for Few-Shot Tuning}

Unlike existing TL autotuning methods, the GC does not benefit from access to extensive or exhaustive datasets.
Without modeling the relationships between performance and parameter configurations, GC lacks a mechanism to disfavor parameter configurations with subpar performance.
Nevertheless, we can intentionally filter training data based on observed high-performance and fit the GC only to high performance configurations.
In the TL setting, GC-based autotuners generate high-performing configurations immediately and minimize the exploration of low-performing regions.

\paragraph{Quantile Filtering}
We investigate dataset filtering based on performance quantiles to include only top-performing configurations in the training data for GC modeling.
The best quantiles should result in a training data subset with similar distribution for high-performing configurations while maintaining ample tuning space coverage.
To motivate the need for the proper threshold quantile, we present a brief analysis from an exhaustively tuned Syr2k task using Kullback--Leibler (KL) divergence~\cite{KLDivergence}, a statistical measure of the difference between two probability distributions.
Zero KL divergence indicates that the compared distributions are identical; increasing differences between compared distributions increases divergence.
We also analyze the tuning space coverage based on the filtered dataset since filtering can prevent some configurations from being generated.

\begin{table}[!h]
    \centering
    \caption{The tuning space coverage and average marginal KL divergence of quantile-based filtering for the Syr2k benchmark. The KL divergence is calculated using the top 10\% of all configurations as a reference, obtained through brute-force.
    }
    {\scriptsize
    \begin{tabular}{|c|c|c|}
    \hline
    Filtering & Tuning Space & KL \\ Quantile (\%) & Coverage & Divergence \\\hline
    100 & 1.00 & 0.1878 \\\hline

    90 & 1.00 & 0.1713 \\\hline
    80 & 1.00 & 0.1609 \\\hline
    70 & 1.00 & 0.1525 \\\hline

    60 & 0.91 & 0.1409 \\\hline
    50 & 0.91 & 0.1212 \\\hline
    40 & 0.91 & 0.1333 \\\hline

    \rowcolor{blue!25} 30 & 0.82 & 0.1713 \\\hline
    20 & 0.07 & 0.2766 \\\hline
    10 & 0.06 & 0.3079 \\\hline

    \end{tabular}
    }
    \label{tab:kl_divergence}
\end{table}

Quantile filtering cannot be too aggressive.
As shown in Table~\ref{tab:kl_divergence}, the tuning space coverage decreases gradually at first but decreases dramatically (i.e., 0.82 to 0.07) when the filtering quantile decreases from 30\% to 20\%. The significant decrease suggests that a majority of parameter options, especially categorical parameters, have been eliminated.
Generating near-optimal configurations for many tasks requires variation within the tuning space, so filtering must not overly restrict coverage.
The table also indicates that reducing the filtering quantile of source data reduces the average marginal KL divergence between the represented data and the top 10\% of all possible configurations of this particular Syr2k tuning task.
Again, the filtering quantile cannot be reduced indefinitely without consequence: an aggressive filter can increase divergence.
We expect the optimum to change between tasks; overreliance on the source task optimum harms the generalization of the $f(c,t)$ relationship.

A smaller divergence for the filtered source increases the likelihood of sampling optimal configurations by redistributing the sampling probability from suboptimal areas of the space to regions that closely resemble known near-optimal configurations.
Based on this trend and our experiments, we recommend using less than 50\% of the original tuning data to exclude low-performing characteristics from prior data.
This excludes many evaluations that prior autotuners made to inform the surrogate model rather than to improve the best-known optimum.
Empirically, we determine that at least the top 15\% of data from related tuning runs is needed to avoid over-specification.
We utilize the top 30\% of prior data in our experiments to ensure adequate information is available for complex tuning tasks without overly harming the tuning space coverage.

\subsection{Model Inference}

The fitted GC model represents learned distributions that can be used for inference, but additional steps must be taken to utilize it as an effective TL autotuner.

\subsubsection{Conditional Sampling} \label{sec:methodology-conditional_sampling}
Quantile filtering increases the likelihood that a sampled configuration from the GC will reproduce optimal traits in new tasks, but this fails to respect the specific tuning needs for different tasks.
Meaningful transfer between tasks requires us to label fitting data with a representation of the task, $t$.
This permits conditional sampling; we specify the condition that every sample indicates a particular task.
Conditional sampling imposes arbitrary constraints on the model during generation, which affect the distributions of unconstrained variables before generating their values.
We explain the mathematical mechanics of conditional sampling for GCs in greater detail \href{https://www.github.com/tlranda/GC_TLA/tree/master/ConditionalSampling}{in our repository}.
Conditional sampling prompts the GC to reconstruct the best-fit distribution it learned for the indicated task; if that task was not observed in prior tuning, the same model mechanics ``recover'' a transferred relationship for the new task.

Conditional sampling is particularly effective for the GC because it identifies and isolates critical information from the model.
Since the GC operates on filtered high-performing source data, conditional sampling generates configurations that are expected to perform well for the transferred task.

\subsubsection{Advantages over Alternative Generative Models}

Other generative models can fit a labeled dataset and generate constrained samples, but the GC has lower latency and yields more usable samples than alternatives.
Table~\ref{tab:inference} demonstrates the inference latency of comparable deep neural-network-based generative methods such as CTGAN and TVAE~\cite{CTGANandTVAE}. GC has the lowest latency of all, which is comparable to purely random sampling. 

\begin{table}[!h]
    \centering
    \caption{The cost of generating 1,000 unique samples using various techniques. Conditional sampling with the GC has latency similar to random sampling but represents learned relationships without ill-conditioned data.}
    {\scriptsize
    \begin{tabular}{|c|c|c|c|}
    \hline
    \multirow{2}{*}{Method} & \multirow{2}{*}{Time (s)} & \multicolumn{2}{c|}{Reject Reason (\%)} \\\cline{3-4}
     & & Repeated & Ill-Conditioned \\\hline
    Random & 0.24 & -- & -- \\\hline
    GC & 0.52 & 62.13\% & 0\% \\\hline
    CTGAN & 1.28 & 7.33\% & 87.87\% \\\hline
    TVAE & 80.77 & 2.25\% & 97.70\% \\\hline
    \end{tabular}
    }
    \label{tab:inference}
\end{table}

The GC's advantage in latency is partially due to the acceptance rate of generated samples, also shown in Table~\ref{tab:inference}.
The separation of joint and marginal models permits the GC to satisfy constraints before generating other values, so only repeated parameter configurations are removed from its generated configurations.
Some other models, such as the CopulaGAN~\cite{copulaGAN_Citation}, can also utilize conditional sampling; however, it can fail to generate any configurations when prompted to produce out-of-distribution data, which is important for transferring tuning to new tasks.

CTGAN and TVAE generate excess samples and then employ filters to discard ill-conditioned data. These methods are computationally inefficient. While relaxing constraints can help reduce their generation latency, it comes at the cost of compromising the quality of the generated data, which  no longer best fits the desired task.
CTGAN and TVAE are not ideal for few-shot transfer learning autotuning scenarios, where both latency and utility are crucial factors to consider.

CopulaGAN, CTGAN, and TVAE are proposed for synthetic data generation and are effective when provided with large amounts of data.
In autotuning, however, we often have limited data.
GC is more effective for these settings because of its computational simplicity.

\subsubsection{Managing Probability of Success} \label{sec:methodology-probability}

The success rate for generative autotuning is subject to randomness, even though the transferred distribution is biased toward values that are expected to be near-optimal.
Therefore, it is crucial to understand the probabilities involved in GC generation to determine whether the technique is appropriate and what evaluation budget is necessary to expect a certain threshold of success.

The GC's autotuning process samples $k$ configurations without replacement from a distribution that spans $|C|$ potential candidates.
Within the distribution are $|I|$ ideal candidates, which are optimal or near-optimal.
Frequently, the top 1\% of evaluations in real-world benchmarks have nearly equivalent performance.
Identifying one or more of these top 1\% candidates within the budgeted $k$ trials is an acceptable goal for few-shot TL autotuning.
The probability that one or more such ideal candidates are selected within $k$ trials is hypergeometric sampling, described by Equation~\ref{eq:hypergeometric}:

\begin{equation}
	P(\# Optimal \ge 1) = \sum_{i=1}^{k} \frac {\binom{|I|}{i}\binom{|C|-|I|}{k-i}} {\binom{|C|}{k}}.
	\label{eq:hypergeometric}
\end{equation}

If we fit all source task data and $|C|$ is the size of the entire configuration space, then sampling the top 1\% of performance within a few shots is unlikely.
Using quantile filtering on the source data for the GC can make some configurations statistically improbable or impossible to generate, eliminating them from the search.
These excluded configurations are expected to be suboptimal because they fail to exhibit characteristics common with known optimal-like data from source task tuning.

Eliminating suboptimal configurations with quantile filtering reduces the size $|C|$.
Recall from Table~\ref{tab:kl_divergence} that the tuning space coverage decreases dramatically after a certain quantile.
The best filtering quantile will minimize KL divergence from the optimal distribution and limit $|C|$ without overspecifying the search space since the latter also contributes to the probability of the few-shot success.
We can determine the reduced $|C|$ from the GC by estimating the number of unique samples generated by the fitted GC.
Nevertheless, we can only measure the resulting change in $|I|$ with exhaustive evaluation, which is needed to quantify the probability of success in Equation~\ref{eq:hypergeometric}. 

The exact reduction in $|I|$ is unknown but can be modeled as a proportion of the eliminated configurations, which represents the opportunity cost of some removed configurations being optimal.
With adjusted $|C|$ and $|I|$, the value of $k$ in Equation~\ref{eq:hypergeometric} can be increased until the probability meets a desired confidence level.
This provides an adequate budget of evaluations $k$ that generates one or more ideal candidates with probability equal to the specified confidence  (e.g., 95\%).
This budget-engineering calculation operates similarly to a convergence guarantee because it permits evaluations of the GC's viability via the size of its budget constraint without performing any empirical evaluations.

\subsection{Addressing Limitations for Autotuning}

Even with our modifications, a few of the known limitations of GC models have limited significance in our intended use case of TL autotuning for source code annotations.

\subsubsection{Underfitting Cross-Variable Dependencies}
The GC expresses codependence between variables using linear correlation, which will underfit complex variable codependencies.
The GC's correlation is expressed between variable pairs, so the number of simultaneously interacting variables is less important than the complexity of dependence between variable pairs.
In most cases for source-code autotuning, annotations are functionally independent of one another or adhere to the linear correlation that the model can express.

\subsubsection{False Ordering and Transitivity for Categories}
The GC's linearized representation of categorical values implies and attempts to leverage a total ordering that may not exist between categories.
This creates transitive relationships that may prove counterproductive for the marginal optimization of categorical data.
One way to counteract this behavior is to utilize binary expansion or one-hot encodings for each category, but this can create many variables when applied to large categories.
Many source code annotations consist of only two values, such as the presence or absence of a \texttt{\#pragma} annotation, which limits the variable to two categories.
Other categorical variables in annotation autotuning are limited to fewer than ten values, which bound the error that marginal kernels must overcome to acceptable degrees.

\subsubsection{Model-Fitting Complexity}
Fitting a GC has cubic time complexity based on the number of variables due to the joint covariance model.
Other TL methods gain a competitive edge when the GC models fifty or more variables, which can make some modifications, such as one-hot encoding, less desirable in practice.
Source code annotations pose some inherent limits on the number of tunable variables due to the decreasing performance significance of additional, non-bottleneck optimization points in an application.
Larger applications require explicit measures, such as importance sampling, to identify the most critical variables to tune.
Our current techniques continue to rely on experts for annotation and can also rely on them to curate an appropriately sized set of variables.

%% file: svg-inkscape/ApproachDiagram_svg-tex.pdf_tex
\begingroup%
  \makeatletter%
  \providecommand\color[2][]{%
    \errmessage{(Inkscape) Color is used for the text in Inkscape, but the package 'color.sty' is not loaded}%
    \renewcommand\color[2][]{}%
  }%
  \providecommand\transparent[1]{%
    \errmessage{(Inkscape) Transparency is used (non-zero) for the text in Inkscape, but the package 'transparent.sty' is not loaded}%
    \renewcommand\transparent[1]{}%
  }%
  \providecommand\rotatebox[2]{#2}%
  \newcommand*\fsize{\dimexpr\f@size pt\relax}%
  \newcommand*\lineheight[1]{\fontsize{\fsize}{#1\fsize}\selectfont}%
  \ifx\svgwidth\undefined%
    \setlength{\unitlength}{406.2492981bp}%
    \ifx\svgscale\undefined%
      \relax%
    \else%
      \setlength{\unitlength}{\unitlength * \real{\svgscale}}%
    \fi%
  \else%
    \setlength{\unitlength}{\svgwidth}%
  \fi%
  \global\let\svgwidth\undefined%
  \global\let\svgscale\undefined%
  \makeatother%
  \begin{picture}(1,0.74170133)%
    \lineheight{1}%
    \setlength\tabcolsep{0pt}%
    \put(0,0){\includegraphics[width=\unitlength,page=1]{svg-inkscape/ApproachDiagram_svg-tex.pdf}}%
  \end{picture}%
\endgroup%

%% file: sections/design.tex
We evaluate our method and several existing techniques in few-shot TL autotuning with a variety of benchmarks empirically evaluated on a real system.

\paragraph{TL Autotuning Benchmarks} \label{sec:design-problems}
We use the Polybench 4.2~\cite{polybench} benchmark suite and several Exascale Computing Project (ECP) proxy mini-applications to evaluate our GC autotuning methodology.
ECP benchmarks are multithreaded, and one (SWLite) is a GPU program. 
The selected applications are based on our ability to define valuable optimizations in our tuning spaces.

Polybench consists of numerical computation kernels extracted from various application domains.
We utilize six of the most complex benchmarks spanning the application domains of linear algebra, image processing, stencils, and data mining: Syr2k, 3mm, Heat-3d, LU, Covariance, and Floyd--Warshall.

The ECP proxy applications represent essential computational kernels from high-performance computing programs, allowing for highly effective performance analyses and tuning without requiring the time-intensive execution of the entire application.
We include four mini-applications---AMG, RSBench, XSBench, and SW4Lite---with different compute-memory access ratios and memory accesses patterns.

We parameterize each kernel with code modifications in performance-critical sections of the benchmark that may improve performance.
These modifications include tile sizes, loop optimization techniques, parallelization and scheduling strategies, data allocation formats, and multiprocess synchronization frequencies.
Table~\ref{tbl:space} shows the number of unique parameters in each experimental benchmark as well as the combinatoric search space size of all possible parameter configurations.
The largest search space has over 5 million potential configurations. The tuning spaces for Polybench and ECP applications are described in greater detail in Tables~\ref{tbl:polybench_params} and~\ref{tbl:exp_params}, respectively.

\paragraph{Source Tasks and Training Dataset} \label{sec:design-source_task}

To form the prior knowledge for TL autotuning, we use offline autotuning through YTOPT~\cite{ytopt} to collect 200 evaluations in each of three non-target tasks: small, medium, and large.
Since 200 evaluations represent <5\% of the search space, any chosen tuner must span performance for maximally effective TL.
All of our YTOPT autotuning is performed by Bayesian optimization with random forests, tuned for 90\% confidence in the 50$^{th}$ quantile of evaluated performance.

\begin{table}[t]
	\begin{center}
		\caption{Tuning spaces for each benchmark alongside the GC's coverage and budget based on the top-30\% of source evaluations. Specific parameters are described in Tables~\ref{tbl:polybench_params} and~\ref{tbl:exp_params}.} \label{tbl:space}
        {\scriptsize
		\begin{tabular}{|l|c|l|l|l|}
			\hline
			Benchmark & \#Params & \# Configurations & GC Coverage & GC Budget \\
			\hline
			3mm & 10 & 376,320 & $\approx$ 2,500 & -- \\
			Covariance & 5 & 5,324 & $\approx$ 110 & -- \\
			Floyd--Warshall & 5 & 5,324 & $\approx$ 1,800 & 15 \\
			Heat3d & 6 & 10,648 & $\approx$ 1,600 & 8 \\
			LU & 5 & 5,324 & $\approx$ 210 & -- \\
			Syr2k & 6 & 10,648 & $\approx$ 800 & 3 \\\hline
			AMG & 9 & 1,180,980 & $\approx$ 108,500 & 5 \\
			RSBench & 9 & 5,196,312 & $\approx$ 316,800 & 3 \\
			XSBench & 8 & 577,368 & $\approx$ 77,500 & 7 \\
			SW4Lite & 8 & 4,752 & $\approx$ 1,800 & 15 \\
			\hline
		\end{tabular}
        }
	\end{center}
\end{table}

\begin{table*}[h]
    \begin{center}
        \caption{Parameters used to tune Polybench Kernels. Values within brackets indicate the options available for an independent parameter, and a list of brackets represents multiple independent parameters.}
        \label{tbl:polybench_params}
        {\scriptsize
        \begin{tabular}{|l|c|c|c|c|c|c|}
        \hline
        Parameter & Values & Covariance & Floyd-Warshall & Heat3d & LU & Syr2k \\\hline
        \multirow{2}{*}{Tile Sizes} & [4-2048], [4-2048], & [4-128], [4-2048], & [4-128], [4-2048], & [4-128], [4-2048], & [4-128], [4-2048], & [4-128], [4-2048], \\
         & [4-2048] & [4-256] & [4-256] & [4-256] & [4-256] & [4-256] \\\hline
        Loop Interchange & [Yes, N/A] & [Yes, N/A] & [Yes, N/A] & [Yes, N/A] & [Yes, N/A] & [Yes, N/A] \\\hline
        Array Packing & [Yes, N/A] $\times$ 6 & [Yes, N/A] & [Yes, N/A] & [Yes, N/A] $\times$ 2 & [Yes, N/A] & [Yes, N/A] $\times$ 2 \\\hline
        \end{tabular}
        }
    \end{center}
\end{table*}

\begin{table*}[h]
    \begin{center}
        \caption{Parameters used to tune ECP mini-applications. 
        }
        \label{tbl:exp_params}
        {\scriptsize
        \begin{tabular}{|l|c|c|c|c|}
        \hline
        Parameter & AMG & RSBench & XSBench & SW4Lite \\\hline
        \multirow{2}{*}{Tile Sizes} & [10-200], [2-256], & \multirow{2}{*}{[2-256], [2-256]} & \multirow{2}{*}{[2-256], [2-256]} & \multirow{2}{*}{--} \\
         & [2-256], [10-200] & & & \\\hline
        \multirow{2}{*}{Optional Parameters} & \multirow{2}{*}{Parallel For} & \multirow{2}{*}{Parallel For} & \multirow{2}{*}{Parallel For} & Parallel For, Nowait, \\
         & & & & MPI\_Barrier \\\hline
        Parallel For Schedule & -- & [100-2000], [10-200] & [10-160] & [dynamic, static] \\\hline
        \multirow{2}{*}{Unrolling Options} & \multirow{2}{*}{[unroll, N/A]} & \multirow{2}{*}{[unroll, N/A]} & \multirow{2}{*}{[unroll, N/A]} & [unroll (6) \footnotemark[2], unroll, \\
         & & & & no-unroll] \\\hline
        \# Threads & [4-8] & [2-256] & [2-256] & [2-256] \\\hline
        \multirow{3}{*}{KMP Affinity} & [compact, scatter, & [compact, scatter, & [compact, scatter, & \multirow{3}{*}{--} \\
         & balanced] & balanced, none, & balanced, none, & \\
         & & disabled, explicit] & disabled, explicit] & \\\hline
        OMP Proc Bind & -- & -- & -- & [close, spread, master] \\\hline
        OMP Places & [core, threads, sockets] & [core, threads, sockets] & [core, threads, sockets] & [core, threads, sockets] \\\hline
        \end{tabular}
        }
    \end{center}
\end{table*}

Table~\ref{tbl:space} summarizes the tuning spaces of source tasks and includes the GC's predicted evaluation budget based on filtered source data.
The prediction is based on the model's capability to identify one or more evaluations in the top 1\% with 95\% confidence, assuming that as much as 5\% of pruned configurations are potentially optimal.
A dash represents an unknown budget, where the overall problem size is reduced to such a degree that it is impossible to predict a budget requirement using Equation~\ref{eq:hypergeometric}.
In this case, the GC's tuning space coverage could fail to include the optimal region if the transfer relationship is poorly informed.
Hence, we cautiously treat indeterminate budgets the same as few-shot TL for techniques that cannot determine their own budget, and we determine how well the GC can perform using the same budget as prior techniques.

\paragraph{Compared Approaches} \label{sec:experiment-approaches}
We evaluate the following autotuning approaches to demonstrate our advantage:

\begin{itemize}
	\item \textbf{Baseline.}
	Parameter values are taken directly from their respective sources; no parameter tuning is performed except compiler flag -O3. 
	
	\item \textbf{Bayesian Optimization.}
	Bayesian optimization (BO) without TL using YTOPT~\cite{ytopt,WK21,WP23}.
	The autotuner utilizes a random forest surrogate model and a hedged Gaussian process to evaluate the expected improvement of proposed configurations.
	
	\item \textbf{GPTune DTLA.}
	GPTune~\cite{GPTune} with DTLA is a state-of-the-art autotuner that is capable of utilizing TL using a neural joint model to combine Gaussian processes representing individual parameters.
	
	\item \textbf{Gaussian Copula (ours).}
	A GC is fit to the top 30\% performing data from source tasks, then conditionally sampled on the target task to generate configurations.
\end{itemize}

\paragraph{Autotuning Procedure} \label{sec:experiment-autotune}
Each benchmark has three source task sizes (small, medium, and large) based on given magnitudes of performance-scaling input features.
The three target task sizes are small-medium (SM), medium-large (ML), and extra-large (XL).
The first two represent two interpolations between source tasks, and the last is an extrapolation beyond the scope of source tasks.
Each target is tuned independently after the model is exposed to all source datasets.
In order to permit the fairest possible comparison among different techniques, the same source tasks dataset is presented to each transfer-capable technique, but the GC filters source datasets for its benefit.
In order to mitigate the variance of randomness employed by each technique, the few-shot tuning process is repeated with three random seeds, and results are reported using the average across all seeds.

We permit each autotuning technique a fixed budget of 30 evaluations per target task.
Even when the GC can predict a viable budget of fewer than 30 evaluations shown in Table~\ref{tbl:space}, we collect all 30 and specifically note the intermediate results when the predicted budget is exhausted.
Since we expect TL techniques to extract some understanding of the problem from prior data, we evaluate success primarily based on the best-observed performance among limited evaluations.

For all evaluations of source and target tasks, the parameterized code is compiled once and executed three times.
The autotuning objective is reported as the mean of the last two evaluations to minimize the impact of variance and uncontrolled noise in the execution environment.
Timing data is collected with internal measurements in the benchmark source code to ensure that overheads such as process startup and data initialization are excluded.

\paragraph{Experimental Platform} \label{sec:design-platform}
All experiments are conducted on a Linux machine with 320 GB 2x AMD EPYC 7742 64-core processor (128 total cores) 1 TB DDR4 with Ubuntu 20.04.2 LTS.
The machine also includes a 40 GB NVIDIA A100, which we use for evaluating the GPU-based SW4Lite ECP application.
Measurements of elapsed time include time for sample generation, source code compilation using the Clang compiler, and program execution.
Each benchmark internally measures empirical performance.

Because the tuning spaces we defined express optimizations through Polly~\cite{polly}, a loop optimizer for LLVM, we use a Clang compiler (version 13.0.0) for compilation.
However, any compiler that supports Polly is suitable for replicating our experiments.
Some Polly optimizations can be applied heuristically based on analysis of the LLVM intermediate representation, while others can be induced by programmer-supplied \texttt{\#pragma} directives in the source code.
Currently, not all code transformations can be specified by directives, such as unroll-and-jam, loop fusion, and code motion.
For this reason, two of our applications (3mm and LU) adopt heuristic optimizations.

%% file: sections/experiments.tex
We separate the presentation of our results between the Polybench and Exascale benchmark suites and identify key successes and limitations of our technique compared with the state-of-the-art approaches.

\footnotetext[2]{Specifically unroll loops by a factor of 6 iterations}

\subsection{Polybench Autotuning} \label{sec:experiment-polybench}

The Polybench benchmarks demonstrate several different behaviors for generative autotuning with the GC, including aggressive space pruning, uncertain optimization signals, and high-confidence benchmarks that represent a best-case scenario for the technique.

\begin{table}[h]
\centering
\caption{Autotuning results after a maximum of 30 evaluations; results are averaged across three repeated tuning attempts with unique seeds.}
{\scriptsize
\begin{tabular}{|c|c|c|c|c|c|c|c|}
\hline
\multirow{3}{*}{} & \multirow{3}{*}{App.} & \multirow{3}{*}{Scale} & \multicolumn{5}{c|}{Peak Speedup (\# Evaluation Discovered)} \\\cline{4-8}
 & & & \multicolumn{3}{c|}{GC} & \multirow{2}{0.05\linewidth}{BO Best} & \multirow{2}{0.08\linewidth}{GPTune Best} \\
 & & & 1$^{st}$ & Budget & Best & & \\
\hline
\multirow{15}{*}{\rotatebox[origin=c]{90}{Polybench Kernels}} & \multirow{3}{*}{3mm} & SM & 5.09 & \cellcolor{blue!25} 5.70 (23) & 5.70 (23) & 3.03 (26) & 5.53 (30) \\
 &  & ML & \cellcolor{blue!25} 5.25 & 5.57 (29) & 5.57 (29) & 3.29 (30) & 5.16 (16) \\
 &  & XL & \cellcolor{blue!25} 27.10 & 33.39 (18) & 33.39 (18) & 20.58 (30) & 18.96 (25) \\\cline{2-8}
 & \multirow{3}{*}{Cov.} & SM & 21.10 & \cellcolor{blue!25} 21.98 (21) & 21.98 (21) & 21.83 (28) & 13.30 (30) \\
 &  & ML & \cellcolor{blue!25} 4.13 & 4.27 (26) & 4.27 (26) & 3.87 (25) & 4.07 (30) \\
 &  & XL & \cellcolor{blue!25} 23.04 & 23.96 (2) & 23.96 (2) & 8.43 (12) & 17.88 (9) \\\cline{2-8}
 & \multirow{3}{*}{Floyd-W.} & SM & 1.01 & 1.02 (17) & 1.02 (17) & \cellcolor{blue!25} 1.02 (20) & 1.01 (26) \\
 &  & ML & \cellcolor{blue!25} 1.02 & 1.02 (1) & 1.02 (1) & 1.01 (25) & 1.01 (3) \\
 &  & XL & 0.99 & 1.00 (29) & 1.00 (29) & 1.01 (16) & \cellcolor{blue!25} 1.01 (20) \\\cline{2-8}
 & \multirow{3}{*}{Heat3d} & SM & 1.83 & 2.03 (5) & 2.06 (18) & 2.21 (15) & \cellcolor{blue!25} 2.30 (28) \\
 &  & ML & 1.89 & 1.89 (1) & 2.06 (10) & \cellcolor{blue!25} 2.12 (25) & 1.80 (6) \\
 &  & XL & 1.50 & \cellcolor{blue!25} 2.92 (2) & 3.09 (18) & 2.16 (13) & 2.75 (29) \\\cline{2-8}
 & \multirow{3}{*}{LU} & SM & \cellcolor{blue!25} 1.16 & 1.18 (25) & 1.18 (25) & 1.12 (30) & 1.11 (19) \\
 &  & ML & 1.15 & \cellcolor{blue!25} 1.20 (24) & 1.20 (24) & 1.17 (26) & 1.19 (5) \\
 &  & XL & \cellcolor{blue!25} 1.00 & 1.00 (3) & 1.00 (3) & 0.98 (13) & 1.00 (29) \\\cline{2-8}
 & \multirow{3}{*}{Syr2k} & SM & 2.06 & \cellcolor{blue!25} 2.90 (2) & 3.32 (18) & 2.34 (12) & 2.41 (11) \\
 &  & ML & 0.80 & \cellcolor{blue!25} 1.17 (2) & 1.22 (16) & 0.93 (29) & 0.85 (30) \\
 &  & XL & \cellcolor{blue!25} 0.95 & 1.09 (2) & 1.09 (2) & 0.42 (23) & 0.85 (26) \\\cline{2-8}
\hline
\multirow{12}{*}{\rotatebox[origin=c]{90}{Exascale Computing Proxies}} & \multirow{3}{*}{AMG} & SM & 0.87 & 0.91 (3) & 0.91 (3) & \cellcolor{blue!25} 0.92 (19) & 0.90 (19) \\
 &  & ML & \cellcolor{blue!25} 0.93 & 0.93 (1) & 0.93 (1) & 0.93 (20) & 0.87 (3) \\
 &  & XL & 0.95 & 0.95 (5) & \cellcolor{blue!25} 0.98 (23) & 0.97 (27) & 0.93 (25) \\\cline{2-8}
 & \multirow{3}{*}{RSBench} & SM & \cellcolor{blue!25} 1.40 & 1.40 (3) & 1.40 (8) & 1.25 (29) & 1.13 (22) \\
 &  & ML & 1.02 & 1.04 (2) & 1.04 (15) & 0.97 (22) & \cellcolor{blue!25} 1.04 (27) \\
 &  & XL & 1.00 & 1.00 (1) & 1.01 (10) & 0.97 (14) & \cellcolor{blue!25} 1.02 (18) \\\cline{2-8}
 & \multirow{3}{*}{XSBench} & SM & 1.20 & 1.20 (7) & \cellcolor{blue!25} 1.21 (28) & 1.17 (24) & 1.21 (24) \\
 &  & ML & 1.05 & 1.06 (4) & 1.06 (4) & 1.04 (6) & \cellcolor{blue!25} 1.07 (5) \\
 &  & XL & 1.01 & 1.02 (5) & 1.03 (24) & 0.99 (6) & \cellcolor{blue!25} 1.05 (5) \\\cline{2-8}
 & \multirow{3}{*}{SW4Lite} & SM & 0.99 & \cellcolor{blue!25} 1.00 (6) & 1.00 (6) & 0.98 (26) & 0.99 (17) \\
 &  & ML & 0.99 & \cellcolor{blue!25} 0.99 (10) & 0.99 (16) & 0.99 (3) & 0.99 (30) \\
 &  & XL & 0.99 & \cellcolor{blue!25} 0.99 (12) & 0.99 (12) & 0.99 (1) & 0.99 (14) \\\cline{2-8}
\hline
\end{tabular}
}
\label{tbl:general_results}
\end{table}

\begin{figure*}[!h]
	\centering
		\includegraphics[width=0.68\columnwidth]{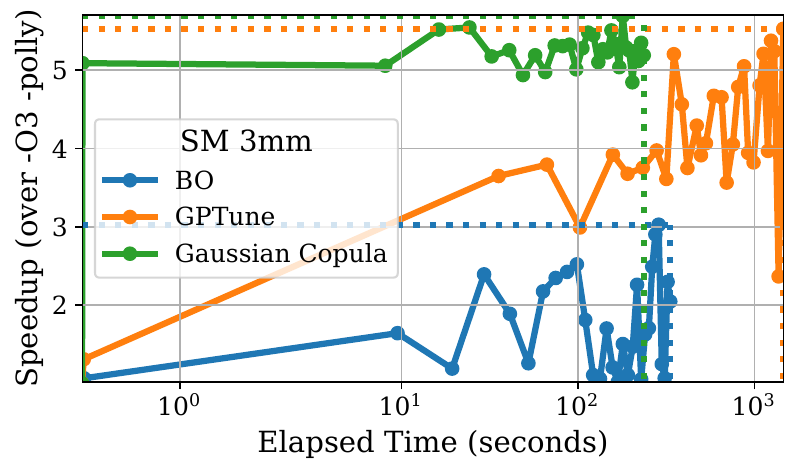}
		\includegraphics[width=0.68\columnwidth]{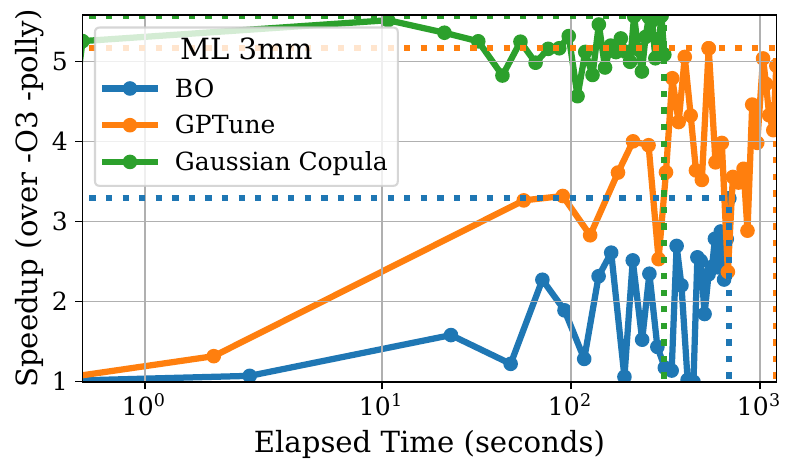}
		\includegraphics[width=0.68\columnwidth]{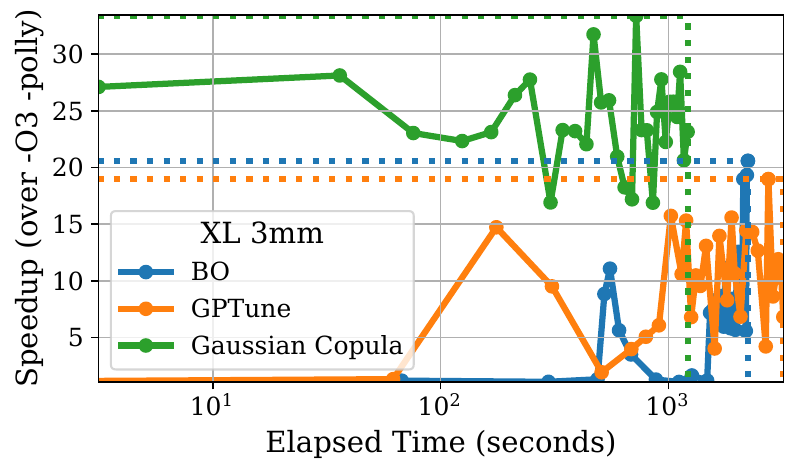}
	\caption{Observed speedup vs. log-scale elapsed time for few-shot TL autotuning. The dotted lines indicate results trimmed to the GC's predicted budget.}
	\label{3mm_and_covariance}
\end{figure*}

General results for the Polybench benchmarks are presented in the upper portion of Table~\ref{tbl:general_results}.
On the 3mm XL task, the GC yields an additional 12.81$\times$ speedup (i.e., 33.39$\times$ vs. 20.58$\times$) compared with prior autotuning techniques.
In half of the Polybench tasks, the GC's first evaluation outperforms the best tuning result discovered by BO or GPTune.
When we utilize the GC's expected budget or the maximum number of evaluations whenever the budget is undefined, the GC outperforms GPTune and BO in over 80\% of all tuning tasks.
Even on tasks where the GC does not outperform prior work, the peak speedup sampled by the GC is within 5.5\% of the peak performance sampled by prior work.

The GC is highly successful both on its first evaluation and within its allotted evaluation budget because of the effectiveness of its search space reductions and distribution transfer through conditional sampling.
Both GPTune and BO must allocate portions of their evaluation budget to explore the space and refine the model's transfer or general surrogate knowledge.
The GC does not need these subpar evaluations.
Thus it can be extremely aggressive in the few-shot tuning, as shown in Figure~\ref{3mm_and_covariance} where nearly every proposed evaluation of the GC outperforms all evaluations proposed by other tuning methods.

\begin{figure*}[!h]
	\centering
		\includegraphics[width=0.68\columnwidth]{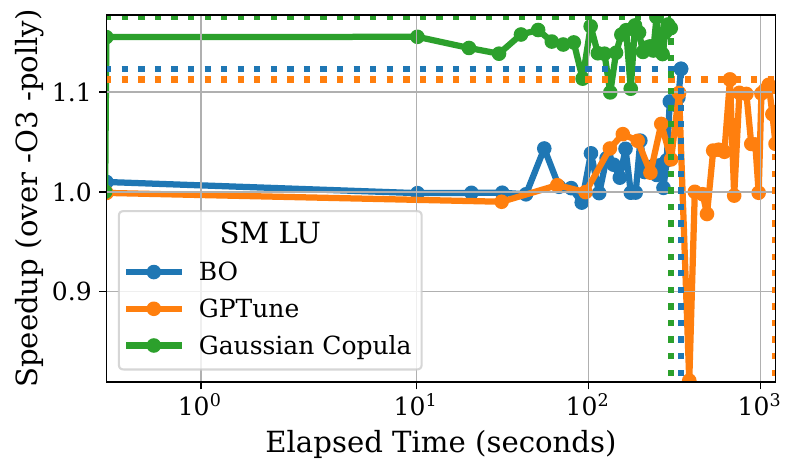}
		\includegraphics[width=0.68\columnwidth]{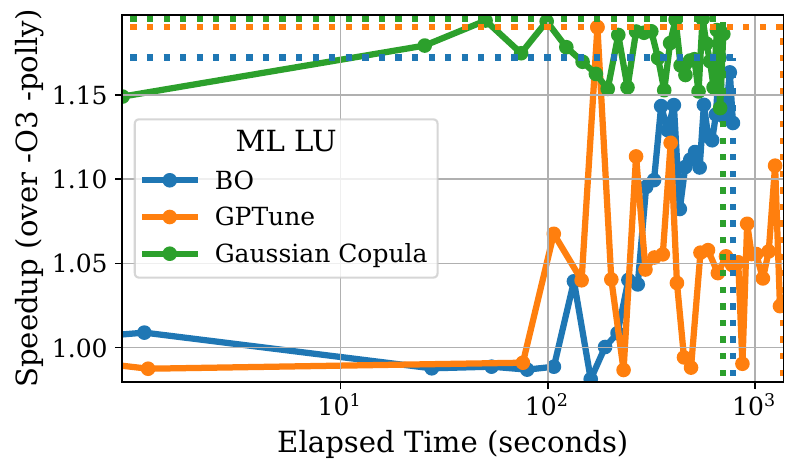}
		\includegraphics[width=0.68\columnwidth]{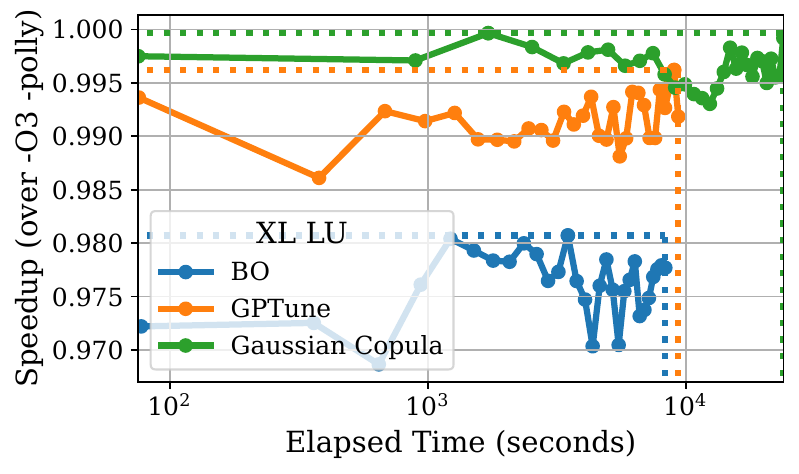}
    \caption{Ambiguous responses to tuning yield minimal speedup, but the GC remains competitive with prior work.}
	\label{fig:FloydWarshall_LU}
\end{figure*}

The GC prunes spaces for 3mm, Covariance, and LU too aggressively for us to predict an evaluation budget.
Our results demonstrate that the GC still identifies the best speedup across all techniques in all tasks for these benchmarks when given the same tuning budget allocated to other techniques.
The search space reduction performed by the GC outperforms prior autotuning by properly identifying characteristics of optimal configurations across tasks and correctly modifying these relationships for each target task.

The Floyd--Warshall and LU benchmarks are challenging for any autotuning technique to optimize.
Without exhaustive data for these benchmarks, it is unclear whether this is due to the original source code parameters being near-optimal or the tuning space exposing mostly unhelpful alterations to the benchmark source.
Critically, the GC still produces highly consistent and comparatively valuable results on each evaluation, as shown for the LU benchmark in Figure~\ref{fig:FloydWarshall_LU}.

\begin{figure}[!h]
	\centering
	\def\svgwidth{.8\linewidth}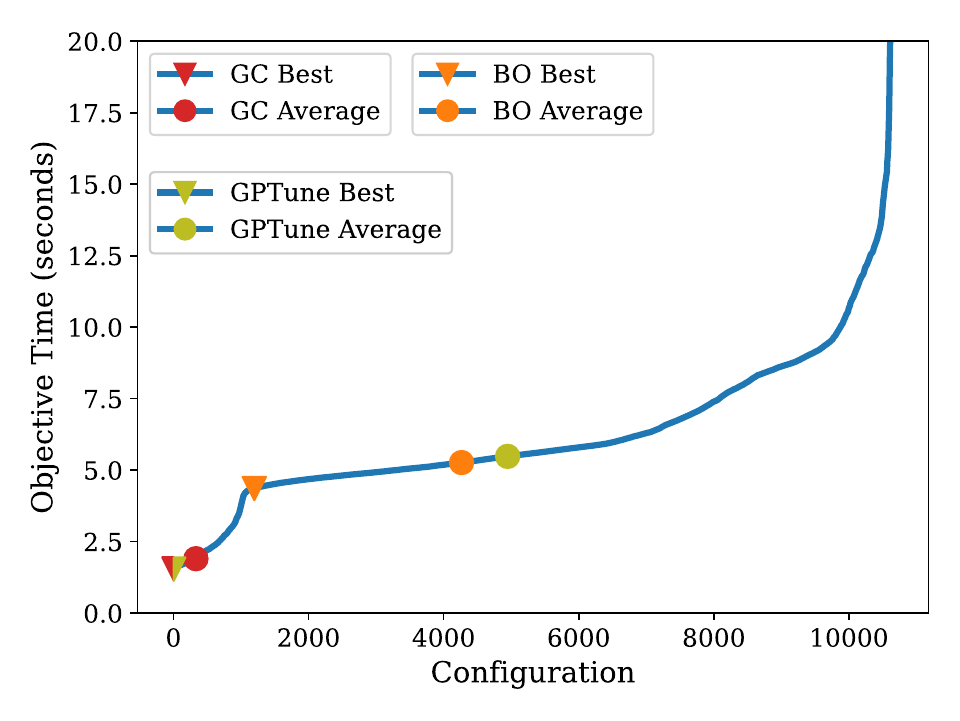
	\caption{Brute-forcing the Syr2k XL task proves that the GC and GPTune can identify the global optimum in 30 evaluations, but the GC avoids poor evaluations, giving it better average performance.}
	\label{fig:exhaustive}
\end{figure}

To ensure that few-shot TL autotuning is effective, we brute-force all configurations of the Syr2k XL task in Figure~\ref{fig:exhaustive}.
Both the GC and GPTune closely approximate the global optimum within a few shots.
However, all evaluations proposed by the GC are near-optimal, while other methods require repeated exploration of poor-performing regions to identify their transfer relationships.

\vspace{-1.5em}
\subsection{Exascale Miniapplications Autotuning} \label{sec:experiment-exascale}

The selected exascale benchmarks represent the most significant challenge for few-shot TL autotuning, with search spaces that are orders of magnitudes larger than those present in the Polybench kernels and complex interplay between many variables.
We expect less speedup from autotuning spaces for these applications for several reasons.
First, the tuning spaces are orders of magnitude larger than Polybench tuning spaces; we use the same number of source task evaluations for all experiments, which means that TL operates on less complete information about each ECP tuning problem.
Second, for more advanced applications, it is more challenging to represent highly effective tunable optimizations than the more straightforward Polybench kernels.
Third, some speedup from system-related tuning parameters can be hidden by other tuning parameters.
The choice of core affinity, for example, has a greater impact on performance if the configuration also includes many threads.
Finally, some parameter defaults, such as loop tiling values, are already highly effective, which limits the improvement that can be extracted from the tuning space.
Although we temper our expected improvement from autotuning, these experiments represent more realistic tuning scenarios where autotuning refines more complex and partially optimized code.

\begin{figure*}[h]
	\centering
	\subfloat[XSBench]{%
		\includegraphics[width=0.68\columnwidth]{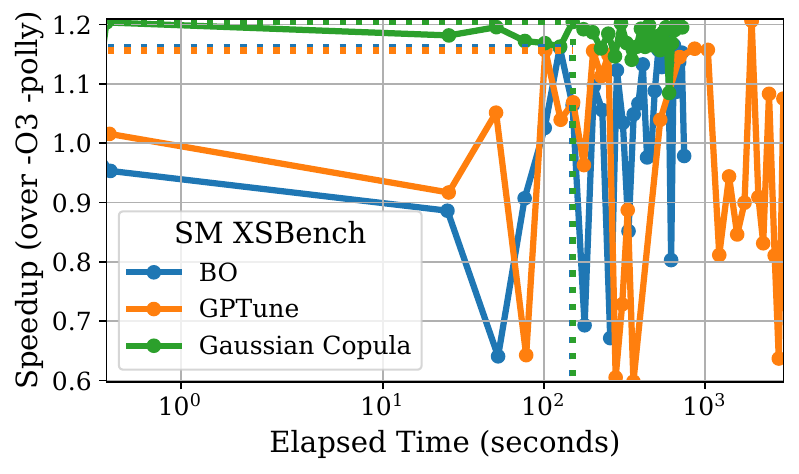}
		\includegraphics[width=0.68\columnwidth]{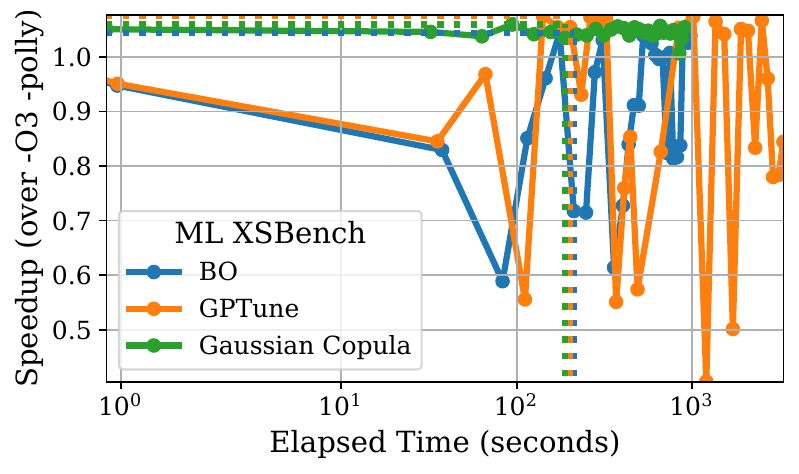}
		\includegraphics[width=0.68\columnwidth]{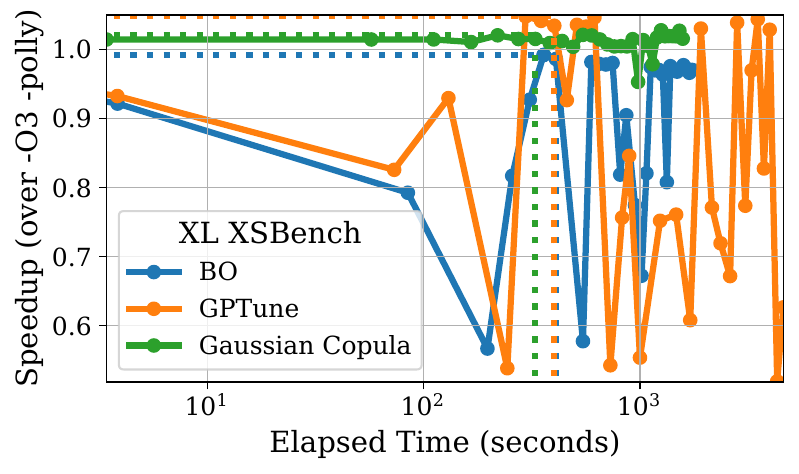}
	}\label{fig:exascale_a}
	\\
	\subfloat[SW4Lite]{%
		\includegraphics[width=0.68\columnwidth]{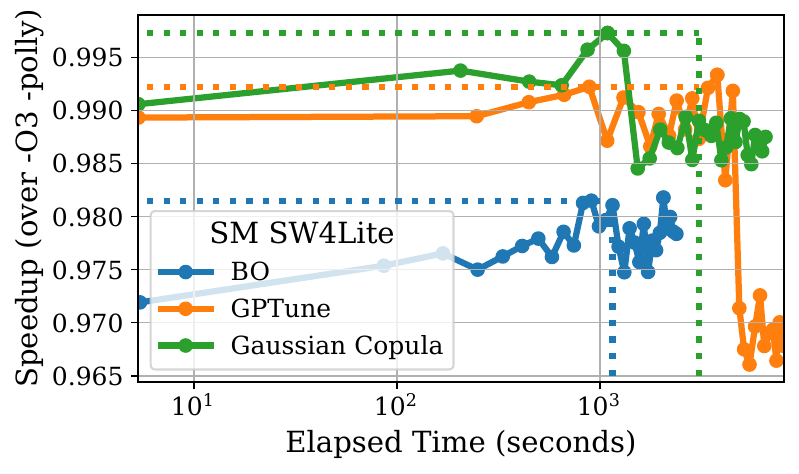}
		\includegraphics[width=0.68\columnwidth]{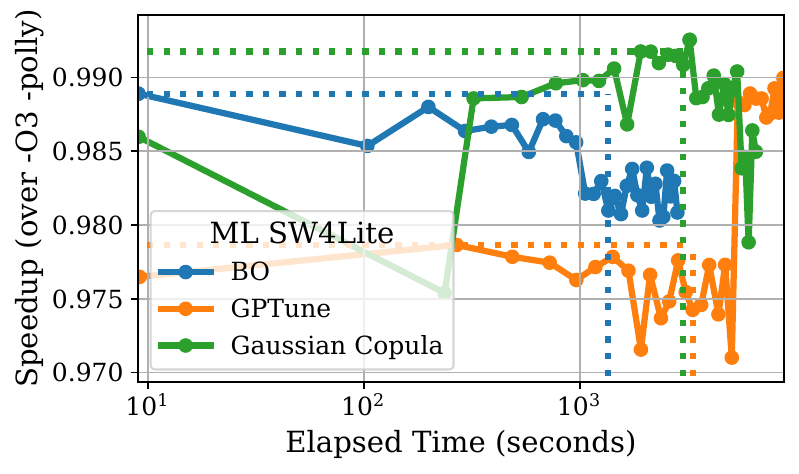}
		\includegraphics[width=0.68\columnwidth]{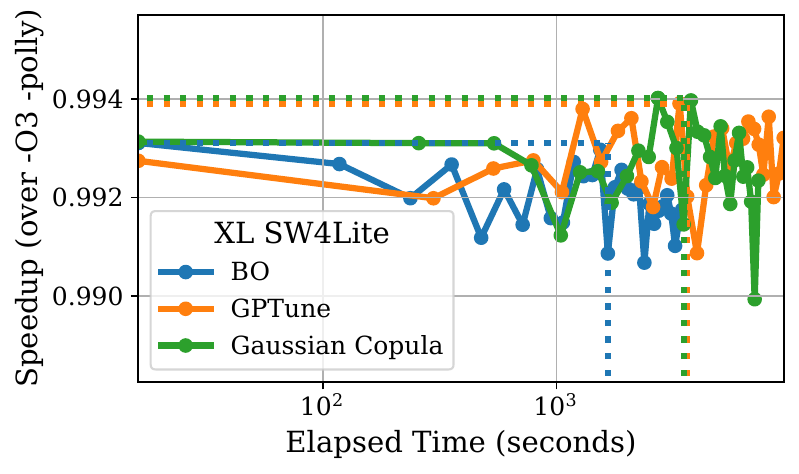}
	}\label{fig:exascale_b}
    \caption{The GC remains competitive with state-of-the-art techniques on complex ECP benchmarks.}
	\label{fig:exascale}
\end{figure*}

Even though our GC technique cannot leverage information gained through iterative evaluations, the technique meets or exceeds the original expert-optimized performance on over half of the exascale tuning tasks.
The AMG task is the most difficult for any technique to optimize, but the GC outperforms GPTune either from its first evaluation or within its predicted budget for all transfer tasks.
Even as the relationship between parameters and performance becomes more complex and search spaces grow orders of magnitude larger, the GC can identify high-performing traits in prior data and produce high-quality candidates in the few-shot tuning scenario.
Across all exascale benchmarks, the GC produces configurations within a performance margin of 2\% of those discovered by GPTune at worst.
Notably, GPTune's best evaluations for two XSBench tuning tasks are better than ours, but the superior evaluations are collected during its random sampling for the new task, as shown in Figure~\ref{fig:exascale}a.
This may indicate that the prior tuning data does not adequately inform autotuning techniques of characteristics of the optimum for this benchmark.

We also note that the GC retains the black-box characteristics enjoyed by prior methods such as BO.
Unlike other benchmarks in this work, SW4Lite is a GPU-enabled benchmark, and the tuned kernel is executed on GPU hardware.
As shown in Figure~\ref{fig:exascale}b, the GC evaluates higher-performing configurations than exploratory techniques such as GPTune do.
The proposed tuning budget is also reliable across multiple seeds, such that the GC reliably makes its best evaluation within the budgeted number of evaluations.
If much larger budgets are allowed, the GC has less chance of improving than other TL autotuning techniques have. 
In such cases, or if any possible performance gain is desired, our technique may be best utilized to perform initial exploration of new spaces within a limited few-shot budget to bootstrap iterative techniques.

%% file: svg-inkscape/BruteForce_svg-tex.pdf_tex
\begingroup%
  \makeatletter%
  \providecommand\color[2][]{%
    \errmessage{(Inkscape) Color is used for the text in Inkscape, but the package 'color.sty' is not loaded}%
    \renewcommand\color[2][]{}%
  }%
  \providecommand\transparent[1]{%
    \errmessage{(Inkscape) Transparency is used (non-zero) for the text in Inkscape, but the package 'transparent.sty' is not loaded}%
    \renewcommand\transparent[1]{}%
  }%
  \providecommand\rotatebox[2]{#2}%
  \newcommand*\fsize{\dimexpr\f@size pt\relax}%
  \newcommand*\lineheight[1]{\fontsize{\fsize}{#1\fsize}\selectfont}%
  \ifx\svgwidth\undefined%
    \setlength{\unitlength}{460.79998779bp}%
    \ifx\svgscale\undefined%
      \relax%
    \else%
      \setlength{\unitlength}{\unitlength * \real{\svgscale}}%
    \fi%
  \else%
    \setlength{\unitlength}{\svgwidth}%
  \fi%
  \global\let\svgwidth\undefined%
  \global\let\svgscale\undefined%
  \makeatother%
  \begin{picture}(1,0.75000003)%
    \lineheight{1}%
    \setlength\tabcolsep{0pt}%
    \put(0,0){\includegraphics[width=\unitlength,page=1]{svg-inkscape/BruteForce_svg-tex.pdf}}%
  \end{picture}%
\endgroup%

%% file: sections/relwork.tex
Prior TL autotuning has enabled data reuse on related tasks for increased sampling efficiency and reduced modeling overhead.
BLISS~\cite{Roy2021} attributes significant cost to tuning multiple models for large-scale applications but also demonstrates that it is difficult to generalize between small datasets and the full range of potential performance.
Other work~\cite{jamshidi2017transfer} employs cost models to substitute cheaper sources of information and utilizes TL to generalize information as needed.
However, in situations with a limited budget, the cost model is less relevant to the target problem and requires model reconstruction. 
Projecting an optimum via machine learning techniques such as GPTune~\cite{sid2019multitask, GPTune} enables more budget optimization for few-shot transfer, but these models require blind evaluations in each new task to form the basis for the transfer relationship.
Other works such as Active Harmony, ANGEL, and ParEGO~\cite{Ananta2011, Chen2015,ParEGO} focus on multiobjective efficiency by refining a surrogate Pareto frontier.
These algorithms provide stronger long-term convergence guarantees rather than few-shot performance.
Our work permits immediate access to the most efficient samples through conditional sampling, allowing for aggressive few-shot tuning.

Prior works have also used biased sample distribution and importance sampling to increase autotuning capabilities.
Marathe et al.~\cite{Marathe2017} found that the correlation between different input scales and available parallelism improves performance predictions.
Their work, however, intends to optimize for common-case average outputs and cannot drive the search aggressively.
GEIST~\cite{Thiagarajan2018} transforms the problem of bias and variance in parameter spaces into undirected graphs and reframes the optimization problem into predicting labels for high or low performance.
The autotuning framework Tuneful~\cite{Tuneful} utilizes incremental sensitivity analysis in BO and explicitly utilizes importance to identify performance trends.
Chen et al.~\cite{EfficientCompilerAutotuningViaBO} use random forest importance measures in massive search spaces by limiting the number of simultaneously tuned parameters to permit full-space exploration.
Our biased generative GC reinforces \textit{and} benefits from increased likelihood to sample the most important parameters of a search space.

Copulas have been reported in the literature as part of an autotuning process.
Salinas et al.~\cite{QuantileHyperparameterTL} used the GC process to bootstrap expected improvement from a small number of initial samples in a BO framework based on ranked quantiles.
More recently, Zhang et al.~\cite{FastParamTuningFrameworkTLMOBO} utilized the correlation identified by a GC to explore the multiobjective Pareto frontier.
Both studies used the GC to aid the BO process in TL.
Salinas et al.~\cite{QuantileHyperparameterTL} used GCs to build an expected improvement autotuning model with minimal initial random samples or prior data and iteratively refit the model as information became available.
The effectiveness of copulas in these techniques is limited to variable correlations in relatively low degrees.

Our work uses the traditional GC with some modifications from the SDV~\cite{Sdv} implementation.
While our experiments do not yield evidence that special care in dependence modeling is necessary, we note that different copulas or GCs are available~\cite{CopulaMarginalDistributions}.
Users can select between variations better suited for tail distributions for which Pearson correlation is insufficient to describe covariant behaviors jointly.

\vspace{-3mm}

%% file: sections/conclusions.tex
In this work, we propose the GC as the first generative TL-based autotuning technique.
Our technique aggressively searches for best-performing configurations in few-shot settings using quantile filtering and conditional sampling to bias distributions learned by the GC model.
We are the first TL-based autotuning work to include expectation of success as a budget-identifying measure to predict few-shot performance.
We then evaluate our technique on various real-world benchmark applications, demonstrating remarkable effectiveness in few-shot TL settings where continued explorations of benchmark characteristics performed by other methods are wasteful resource expenditures. 

Many avenues remain for further research with this generative TL-based autotuning framework, including multi-node evaluations, multiobjective tuning, and modifications to the internal state that permit iterative refinement or bootstrapping for continued tuning similar to the prior work.